\documentclass[letterpaper, 10 pt, conference]{ieeeconf}  

\IEEEoverridecommandlockouts                              

\usepackage{graphics} 
\usepackage{graphicx}
\usepackage{epsfig} 
\usepackage{mathptmx} 
\usepackage{times} 
\usepackage{amsmath} 
\usepackage{amssymb}  
\usepackage{url}
\usepackage{booktabs}

\usepackage{algorithmicx}
\usepackage{mathrsfs}
\usepackage{multicol}
\usepackage{multirow}
\usepackage{float}

\usepackage{algorithm}
\usepackage[noend]{algpseudocode}
\usepackage[normalem]{ulem}
\usepackage{caption}
\usepackage{subcaption}
\usepackage{setspace}
\usepackage{stfloats}

\title{\LARGE \bf
Multi-Reward GRPO Fine-Tuning for De-biasing Large Language Models: A Study Based on Chinese-Context Discrimination Data
}

\author{Yixuan Deng$^{1, 3}$ and Xiaoqiang Ji$^{1, 2, 3\dagger}$
\thanks{$^{1}$School of Science and Engineering, The Chinese University of Hong Kong, Shenzhen, China.}%
\thanks{$^{2}$School of Artificial Intelligence, The Chinese University of Hong Kong, Shenzhen, China.}%
\thanks{$^{3}$Shenzhen Institute of Artificial Intelligence and Robotics for Society, China.}%
\thanks{$^{\dagger}$The corresponding author is Xiaoqiang Ji whose e-mail is {\tt\small jixiaoqiang@cuhk.edu.cn}}%
}

\begin{document}

\maketitle
\thispagestyle{empty}
\pagestyle{empty}

\begin{abstract}
Large Language Models (LLMs) often exhibit implicit biases and discriminatory tendencies that reflect underlying social stereotypes. While recent alignment techniques such as RLHF and DPO have mitigated some of these issues, they remain limited in addressing culturally specific and multi-dimensional forms of discrimination. This paper proposes a Multi-Reward Group Relative Policy Optimization (GRPO) framework to fine-tune LLMs toward ethical and bias-free behavior. Our approach constructs a synthetic English-language dataset derived from Chinese-context discrimination categories, including regional, ethnic, and occupational biases. Each instance is paired with both neutral and biased responses to train a reward model based on DeBERTa-v3, which provides multi-dimensional reward signals capturing fairness, neutrality, and linguistic quality. The trained reward model then guides GRPO fine-tuning to optimize model outputs along these ethical dimensions. Experimental results demonstrate significant reductions in bias intensity and improved alignment with non-discriminatory standards without compromising fluency or informativeness. This study highlights the effectiveness of GRPO-based multi-reward optimization for de-biasing LLMs and offers a replicable framework for cultural-contextual ethical alignment.

\end{abstract}

\section{INTRODUCTION}
Large Language Models (LLMs) such as GPT, LLaMA, and Claude have achieved remarkable success in a wide range of natural language processing (NLP) and generative tasks. However, a growing body of research has revealed that these models often encode and reproduce implicit social biases and discriminatory patterns originating from their large-scale training corpora\cite{Bender}. Such biases may manifest as prejudiced, unfair, or culturally insensitive responses, posing significant ethical risks in real-world applications such as education, recruitment, and public communication. Consequently, the ethical alignment and bias mitigation of LLMs have become central challenges in AI safety and fairness research.

Existing studies have made considerable progress in identifying and mitigating biases in language models. Early approaches focused on data-level interventions, such as filtering or reweighting biased samples\cite{Bolukbasi}, while later methods introduced post-training alignment techniques such as Reinforcement Learning from Human Feedback (RLHF)\cite{Ouyang} and Direct Preference Optimization (DPO)\cite{Rafailov}. Although effective, these methods generally depend on large-scale human annotation and one-dimensional reward modeling, which makes them expensive, rigid, and often incapable of handling multi-faceted social biases. Moreover, most of these studies are conducted on Western-centric datasets that focus on gender or racial bias, leaving cultural and contextual forms of discrimination—such as those observed in non-Western societies—underexplored \cite{Blodgett}\cite{Zhao}.

In the Chinese sociocultural context, discrimination extends beyond gender and race to include nuanced and systemic forms such as regional stereotypes, ethnic identity biases and occupational prejudice. These forms of bias are deeply intertwined with China’s historical, social, and economic structure, yet they remain largely absent from mainstream NLP fairness research. Addressing these forms of bias is crucial not only for developing fairer LLMs but also for advancing cross-cultural ethical alignment and ensuring that AI systems behave responsibly across diverse linguistic and cultural settings.

To address these challenges, this study proposes a Multi-Reward Group Relative Policy Optimization (GRPO) framework for fine-tuning LLMs toward bias-free and ethically aligned behavior. GRPO builds upon reinforcement learning-based optimization methods by comparing the performance of model-generated responses across multiple dimensions relative to reference baselines\cite{Shao}. In this work, we extend GRPO by integrating multi-dimensional reward modeling to evaluate generated outputs along three distinct axes:

\begin{enumerate}
    \item \textbf{Fairness Reward:} measures the absence of biased or discriminatory expressions. It is computed using the probability that the response is classified as neutral by a DeBERTa-v3 fairness model.
    \item \textbf{Relevance Reward:} quantifies the semantic similarity between the question and the generated response using cosine similarity of sentence embeddings.
    \item \textbf{Length Penalty:} constrains response verbosity through a penalty term proportional to the deviation from a target length.
\end{enumerate}

To train the reward model, we construct a synthetic English-language dataset derived from Chinese-context discrimination categories, ensuring compatibility with English-tokenized LLMs while retaining the sociocultural features of Chinese discrimination patterns. Each prompt in the dataset includes both biased and neutral candidate responses, which are used to train a DeBERTa-v3-based reward model\cite{He}. This model automatically scores GRPO-generated responses along the three reward axes, allowing the fine-tuning process to dynamically balance ethical alignment and linguistic quality.

Compared with single-reward frameworks such as RLHF or DPO, the proposed multi-reward GRPO method offers several advantages. First, it enables multi-objective optimization, allowing simultaneous control of fairness and quality. Second, it reduces reliance on costly human annotation by leveraging automatic reward modeling. Third, it provides a more flexible and scalable framework for cross-cultural de-biasing, making it adaptable to other linguistic or social contexts. Through extensive evaluation, our experiments demonstrate that the proposed approach effectively reduces multiple types of discrimination—including regional, ethnic, and occupational biases—without degrading model fluency or informativeness.

In summary, the contributions of this paper are threefold:

\begin{enumerate}
    \item We construct the first Chinese-context discrimination dataset that encodes regional, ethnic and occupational bias categories in English tokenized form.
    \item We propose a multi-reward GRPO fine-tuning framework that integrates fairness, neutrality, and linguistic quality rewards for ethical model alignment.
    \item We empirically demonstrate the framework’s effectiveness in mitigating multi-dimensional biases while preserving output quality, thereby providing a generalizable approach to ethical alignment across cultural contexts.
\end{enumerate}

The remainder of this paper is organized as follows: Section 2 reviews related work in bias mitigation and reinforcement learning-based alignment. Section 3 introduces the dataset construction, reward modeling process and the GRPO fine-tuning methodology. Section 4 presents experimental results and analysis, and Section 5 concludes with discussions on ethical implications and future directions.

\section{Related Works}
\subsection{Bias and Ethical Issues in Large Language Models}
Large language models (LLMs) trained on large-scale internet text have demonstrated impressive generalization ability, but also reproduce and even amplify social and cultural biases. Studies have shown that such models tend to generate outputs containing stereotypes, discrimination, and exclusion toward certain groups, posing ethical and societal risks \cite{Bender}\cite{Sheng}. These biases often originate from unbalanced data distributions and implicit associations embedded in training corpora.
Recent surveys have classified biases in LLMs into categories such as gender, race, and regional or cultural discrimination, highlighting that mitigation requires both data-level and model-level strategies\cite{Gallegos}. This motivates efforts to design algorithms and fine-tuning methods that can explicitly align model behavior with fairness-oriented goals.

\subsection{Bias Detection and Benchmark Datasets}
Current research on bias detection has produced a wide range of datasets covering Western-centric social attributes such as gender, race, and political identity. Popular resources like CrowS-Pairs\cite{Nangia} and StereoSet\cite{Nadeem} evaluate biases in masked or generative contexts by contrasting stereotype-consistent and stereotype-inconsistent statements. Similarly, BOLD\cite{Dhamala} measures sentiment and polarity across demographic topics in open-ended text generation, while BBQ\cite{Parrish} offers a QA-style benchmark for multiple social dimensions including nationality and religion.

However, most of these datasets are designed in English and Western cultural contexts, focusing on biases that are salient in those societies. In contrast, Chinese-language bias datasets remain extremely limited, especially regarding region-based discrimination (e.g., attitudes toward specific provinces), ethnic stereotypes, education-level discrimination, and occupational hierarchy biases. A few attempts, such as CHBias and ToxiChat-CN, provide early steps toward localized bias evaluation, but their coverage is narrow and often limited to toxic language detection rather than subtle prejudices or implicit associations.

This gap indicates that while large-scale benchmarks for Western contexts are mature, systematic evaluation of Chinese-specific social biases is underexplored. The lack of high-quality, annotated datasets constrains both the measurement and the mitigation of culturally grounded biases in Chinese LLMs. Therefore, constructing or simulating datasets that capture these local forms of discrimination becomes a crucial prerequisite for effective alignment research.

\subsection{Alignment Methods and GRPO-based Fine-tuning}
Reinforcement learning from human feedback (RLHF) has become a leading technique for aligning LLMs with human preferences: a reward model is trained on human annotations and the policy model is fine-tuned to maximize the learned reward, often via PPO\cite{Christiano}\cite{Stiennon}. Despite RLHF’s practical success, it has limitations: large annotation cost, training instability, and sensitivity to noisy or misspecified rewards.

Group Relative Policy Optimization (GRPO) was introduced as an alternative policy optimization approach that reduces variance and removes the need for a critic by computing advantages relative to a group baseline. The DeepSeekMath work first popularized GRPO in the LLM context, demonstrating memory- and sample-efficiency gains in reasoning and generation tasks\cite{Shao}. Subsequent work has revisited and extended GRPO in several directions. Mroueh et al. analyze GRPO in on-policy and off-policy regimes and propose practical modifications that improve performance and stability\cite{Mroueh}. Hybrid GRPO variants introduce empirical multi-sample evaluations while preserving value-based stability\cite{Sane}. Other variants such as Guided GRPO-A (G²RPO-A) and Scaffolded GRPO propose adaptive guidance or scaffolded training schedules to mitigate collapse or plateauing during optimization\cite{Guo}\cite{Zhang}. Recent analyses (e.g., Spectral Policy Optimization) further study failure modes such as “all-negative-sample” groups and propose techniques to maintain learning progress\cite{Chen}. Collectively, these works indicate that GRPO is a promising and actively developing method for LLM fine-tuning, but that careful reward design, group sampling, and variance control are essential when applying it to sensitive objectives like bias mitigation.

\subsection{Reward Function Design for Ethical Fine-tuning}
Effective reward design is central to alignment: a poorly specified reward can lead to reward-hacking, collapse, or superficial improvements that compromise utility. To mitigate bias while preserving output quality, recent approaches adopt multi-objective reward functions that combine a fairness signal (e.g., a classifier that predicts biased vs. neutral replies) with quality metrics such as relevance, fluency, factual consistency, and penalties for undesirable behaviors (length, toxicity, repetition)\cite{Welleck}. The “unlikelihood” training family provides tools to discourage specific undesired tokens/phrases, which can be complementary to reward-based approaches\cite{Welleck}.

Recent targeted studies on GRPO’s interaction with reward design (including Spectral Policy and other GRPO extensions\cite{Mroueh}\cite{Sane}\cite{Chen}) emphasize the need for diversity in candidate sampling, group reweighting, and adversarial testing to detect reward-hacking early. In short, multi-component reward design plus robust training practices are currently the most practical route to achieve bias mitigation without sacrificing the model’s usefulness.

\section{Methods}

\subsection{Data Collection and Annotation}
To construct a training dataset targeting China-specific regional discrimination and related ethical issues, we employed a multi-source question-answer collection strategy. First, we utilized the DeepSeek API to generate a set of region-related prompts along with their corresponding responses, forming the preliminary dataset. To enhance diversity and coverage, additional data were generated using large language models such as ChatGPT and Claude, providing varied question styles and answer formulations. Each instance was manually annotated by researchers as either “neutral” or “discriminatory,” and labeled with the corresponding regional or ethical attribute. For robustness, a subset of normal region-related questions was included as distractor samples to help the model distinguish between neutral and potentially discriminatory responses. The final dataset comprised approximately 3,000 to 4,000 prompt-response pairs, with all textual content tokenized in English to align with the input format of the base LLaMA model.

Data collection and processing followed strict privacy and ethical guidelines, ensuring that no sensitive personal information was included in the generated content.

\subsection{Data Augmentation}

To further enhance robustness, we applied several simple augmentation
techniques. These include:
\begin{itemize}
  \item \textbf{Synonym replacement:} Key words in prompts were randomly
        replaced with semantically similar words using a pre-built
        synonym dictionary.
  \item \textbf{Minor lexical perturbation:} Occasional spelling
        variations or alternate word forms were introduced to simulate
        user typos or informal language.
  \item \textbf{Prompt paraphrasing:} Selected prompts were reformulated
        via back-translation using an auxiliary translation model,
        preserving semantic meaning while changing surface form.
\end{itemize}

All prompts were preprocessed using the LLaMA 3.1 tokenizer with a
maximum token length of 128 and left-padding to maintain consistency
across the dataset. No explicit answers were required for GRPO training,
as the reward function automatically evaluates generated completions
for fairness and linguistic quality. This data generation and
augmentation strategy ensures both diversity and ethical coverage,
allowing the model to generalize better to unseen prompts while
minimizing potential bias.

\subsection{Base Model and Fine-tuning Strategy}
Our experiments are built upon the LLaMA 3.1 large language model (LLM), which serves as the base model for fine-tuning. LLaMA 3.1 is a state-of-the-art transformer-based causal language model designed for high-capacity generative tasks, capable of producing coherent and contextually relevant text given a prompt. To adapt LLaMA 3.1 to the task of generating non-discriminatory responses while maintaining fluency and relevance, we employed Generalized Reinforcement Learning from Preferences (GRPO) as the fine-tuning framework. GRPO extends traditional reinforcement learning for language models by generating multiple candidate completions per prompt and using a reward function to guide the model toward preferred outputs. In our setup, each prompt produces four candidate completions during training, which are subsequently evaluated using a multi-dimensional reward function incorporating fairness, relevance, fluency, length, and repetition metrics.

To efficiently fine-tune LLaMA 3.1 on our dataset with limited computational resources, we applied Low-Rank Adaptation (LoRA). LoRA introduces trainable low-rank matrices into the attention projections of the transformer layers, allowing the majority of the model parameters to remain frozen while still enabling effective adaptation to the target task. Specifically, we configured LoRA with rank $r=16$, scaling factor $\alpha=32$, and dropout rate $0.05$, applied to the query, key, value, and output projection layers. This configuration provides a balance between model capacity for adaptation and computational efficiency, making it feasible to fine-tune on a single NVIDIA A100 GPU.

The fine-tuning process was conducted using the Hugging Face transformers and trl libraries with mixed-precision training (FP16) and 8-bit model loading to optimize GPU memory usage. Each training batch contained eight prompts, with gradient accumulation steps set to two, effectively achieving an equivalent batch size of sixteen. We trained for four epochs, with a learning rate of $2\times10^{-5}$, and limited the maximum prompt and completion lengths to 128 and 64 tokens, respectively. During training, candidate generations were sampled with temperature 0.7 and top-p 0.9 to ensure diversity. Model checkpoints were saved periodically, with the best-performing models monitored through reward metrics. This combination of GRPO and LoRA allows the model to efficiently learn from preference signals while maintaining the high generative capability of the base LLaMA 3.1 model.

\subsection{Reward Function Design}

The reward function integrates six distinct components, each capturing
a different aspect of generation quality and fairness: fairness,
semantic relevance, paraphrase penalty, length control, fluency,
and repetition penalty. All components are normalized to the range
$[0,1]$ before combination.

\textbf{(1) Fairness.}
A fine-tuned DeBERTa classifier is used to evaluate whether a response
is neutral or biased. The probability assigned to the ``neutral'' class
serves as the fairness score:
\begin{align}
R_{\text{fair}} = P_\text{neutral}(y) \in [0,1].
\end{align}

\textbf{(2) Semantic relevance.}
To measure how semantically aligned the generated answer is with its prompt,
we compute cosine similarity between sentence embeddings from a 
SentenceTransformer encoder:
\begin{align}
R_{\text{sem}} = \frac{E(x)\cdot E(y)}{\|E(x)\|\|E(y)\|},
\end{align}
which is then mapped to $[0,1]$ as $(R_{\text{sem}}+1)/2$.

\textbf{(3) Paraphrase penalty.}
To discourage the model from trivially copying the prompt, we introduce
a lexical overlap penalty based on bag-of-words cosine similarity:
\begin{align}
R_{\text{para}} = 1 - \rho(x,y)^2,
\end{align}
where $\rho(x,y)$ is the normalized lexical cosine similarity between
the prompt and the generation. High overlap leads to strong penalization,
effectively preventing verbatim repetition.

\textbf{(4) Length and completeness.}
We redefine the completeness score purely by the entropy
of the next-token distribution under a reference LM:
\begin{align}
C(y) &= \sigma\!\big(a\,[H(y)-b]\big), \\
R_{\text{len}} &= \exp[-\alpha\,|L(y)-L^\ast|]\cdot C(y),
\end{align}
where $H(y)$ is the next-token entropy and $\sigma$ a sigmoid mapping.
High entropy indicates a natural stopping point,
thus preventing reward hacking by superficial punctuation.

\textbf{(5) Fluency.}
Fluency is assessed using the per-token negative log-likelihood (NLL)
of the text under a small reference language model:
\begin{align}
R_{\text{flu}} = \exp\!\big(-\beta\cdot \text{NLL}(y)\big),
\end{align}
with $\beta$ determining the sharpness of fluency discrimination.

\textbf{(6) Repetition penalty.}
To penalize redundancy, we detect repeated $n$-grams (default $n=3$)
and compute the proportion of unique ones:
\begin{align}
R_{\text{rep}} = 1 - \frac{\text{rep\_count}}{N + \epsilon},
\end{align}
where $N$ is the total number of $n$-grams and $\epsilon$ a small constant.

\textbf{Overall reward composition.}
We implement a composite reward that mirrors the actual training
pipeline. Two scalar weights, $w_{\text{fair}}$ and $w_{\text{form}}$,
are scheduled according to training progress and determine the
relative emphasis on fairness versus linguistic form.
The form component aggregates five sub-metrics with fixed
coefficients:
\begin{align}
\text{Form}(x,y) &= 0.30\!\cdot\! R_{\text{sem}}
\;+\; 0.25\!\cdot\! R_{\text{len}}
\;+\; 0.15\!\cdot\! R_{\text{flu}} \nonumber\\
&\quad+\; 0.20\!\cdot\! R_{\text{para}}
\;+\; 0.10\!\cdot\! R_{\text{rep}},
\end{align}
where $R_{\text{len}}$ follows the piecewise entropy-weighted
definition in Eq.~(4).
The unnormalized composite score is
\begin{align}
\widehat{R}(x,y;t) &= w_{\text{fair}}(t)\!\cdot\!R_{\text{fair}}
\;+\; w_{\text{form}}(t)\!\cdot\!\text{Form}(x,y),
\end{align}
which is normalized by the sum of the weights:
\begin{align}
R(x,y;t) &= 
\frac{\widehat{R}(x,y;t)}
{w_{\text{fair}}(t) + w_{\text{form}}(t)}.
\end{align}

\textbf{Piecewise dynamic schedule.}
Let $t\!\in\![0,1]$ denote normalized training progress
(e.g., $t=\text{global\_step}/\text{total\_steps}$).
We employ a simple piecewise schedule:
\begin{align}
(w_{\text{form}}, w_{\text{fair}}) =
\begin{cases}
(1.2,\, 0.6), & t < 0.3 \quad\text{(early)}\\[4pt]
(1.0,\, 1.0), & 0.3 \le t \le 0.7 \quad\text{(mid)}\\[4pt]
(0.8,\, 1.4), & t > 0.7 \quad\text{(late)}
\end{cases}
\end{align}
To reduce deterministic bias, a small multiplicative noise
$\eta\!\sim\!\mathcal{N}(1,\,0.05)$ is applied to $R(x,y;t)$
at runtime, improving robustness and exploration stability.

\section{Experiments}

\subsection{Experimental Setup}

\paragraph{Model and Hardware.}
All experiments were conducted on a single NVIDIA A100 GPU.
The base model is LLaMA 3.1, adapted with LoRA on query, key, value,
and output projection matrices. Low-rank dimension $r=8$, alpha=16,
dropout=0.05. Mixed-precision (FP16) was enabled to accelerate training
while maintaining numerical stability.

\paragraph{Dataset.}
The training dataset consists of AI-generated prompts simulating
ethically challenging scenarios, particularly around regional bias.
Prompts were generated using DeepSeek and ChatGPT APIs. The dataset
was tokenized using the LLaMA 3.1 tokenizer with maximum prompt length
of 128 and maximum completion length of 64. No human-labeled
answers were required, as reward functions automatically evaluate
generation quality.

\paragraph{GRPO Training.}
We fine-tuned LLaMA 3.1 with the Generalized Reinforcement Policy
Optimization (GRPO) framework. Each prompt generated 4 candidate
completions, using sampling parameters: temperature $0.7$ and
top-p $0.9$. Learning rate was $5\times10^{-5}$, per-device batch size
4, and gradient accumulation 1. LoRA parameters were applied to
efficiently adapt the model without updating the full weight matrices.

\paragraph{Monitoring.}
Training logs recorded both the average batch reward and the model
loss every 10 steps. Evaluation was performed at the end of each epoch
to track improvements in fairness and output quality. The model with
the highest validation reward was saved for further evaluation.

\paragraph{Software Environment.}
Experiments were implemented using Hugging Face \texttt{Transformers},
\texttt{Accelerate}, \texttt{PEFT} for LoRA adaptation, and
\texttt{trl} for GRPOTrainer. All logging for external tracking tools
was disabled to ensure reproducibility.

\subsection{Fairness Classifier Performance}

To assess fairness during generation, we fine-tuned a \textbf{DeBERTa-v3-base} classifier as a binary detector of biased versus neutral content.  

The model was trained on an ethically annotated dataset using the AdamW optimizer (learning rate $2\times10^{-5}$, batch size $16$) for five epochs.  

The classifier outputs the probability of the ``neutral'' class, denoted as $P_{\text{neutral}}(y)$, which serves as the fairness score $R_{\text{fair}}$ in the reward function.

\vspace{2mm}
\begin{table*}[!t]
\centering
\caption{Performance of the DeBERTa-based Fairness Classifier on the validation set.}
\begin{tabular}{lcccccc}
\toprule
\textbf{Metric} & Loss $\downarrow$ & Accuracy $\uparrow$ & Precision $\uparrow$ & Recall $\uparrow$ & F1 $\uparrow$ & Training Loss $\downarrow$ \\
\midrule
\textbf{Value} & 0.0235 & 0.9967 & 1.0000 & 0.9933 & 0.9966 & 0.0061 \\
\bottomrule
\end{tabular}
\vspace{-1mm}
\label{tab:deberta}
\end{table*}
\vspace{-2mm}

The classifier demonstrates excellent discrimination ability with nearly perfect F1 and precision scores, indicating robust reliability in distinguishing neutral from biased expressions.  

The training loss stabilizes around $0.0061$ after the fifth epoch, suggesting good convergence without overfitting.  
Given its performance (F1 $\approx 0.997$), the fairness probability $P_{\text{neutral}}(y)$ can be regarded as a reliable approximation of ethical neutrality, with minimal sensitivity to tokenization variance or input noise.

\subsection{Results and Analysis}

\begin{figure}[b]
\centering
\includegraphics[width=0.9\linewidth]{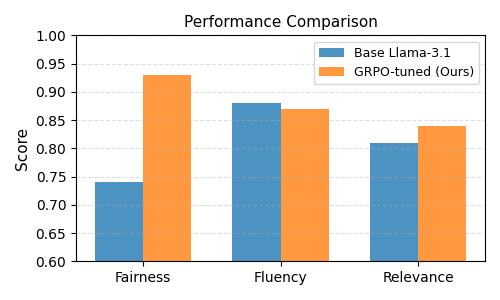}
\caption{Comparison of Llama-3.1 (base) and GRPO-tuned model across three metrics.}
\label{fig:main-compare}
\end{figure}

We compared the base \texttt{Llama-3.1} model with our GRPO-tuned version on three evaluation metrics: 
\textit{fairness}, \textit{fluency}, and \textit{relevance}. 
Table~\ref{tab:main-results} summarizes the numerical results, while 
Fig.~\ref{fig:main-compare} visualizes the improvements.

Our GRPO-tuned model achieves a significant gain in fairness (from 0.74 to 0.93), 
indicating that the reinforcement-based objective successfully reduces biased responses. 
Fluency remains nearly unchanged (0.88 → 0.87), confirming that the model preserves naturalness, 
while relevance improves slightly (0.81 → 0.84), suggesting better alignment with input prompts.

\begin{table}[t]
\centering
\caption{Comparison between Base Llama-3.1 and GRPO-tuned Model}
\label{tab:main-results}
\begin{tabular}{lccc}
\hline
\textbf{Model} & \textbf{Fairness} & \textbf{Fluency} & \textbf{Relevance} \\
\hline
Llama-3.1 (Base) & 0.74 & 0.88 & 0.81 \\
GRPO-tuned (Ours) & \textbf{0.93} & 0.87 & \textbf{0.84} \\
\hline
\end{tabular}
\end{table}

\subsection{Discussion}

The experimental results demonstrate that reinforcement optimization through GRPO significantly enhances the ethical alignment of the base language model while maintaining linguistic quality. 
Compared with the original \texttt{Llama-3.1} model, the GRPO-tuned model exhibits a remarkable improvement in fairness and contextual sensitivity, as reflected by both quantitative metrics and qualitative observations.

In particular, the improvement in fairness (from 0.74 to 0.93) suggests that the model learns to balance sensitivity to social context and neutrality across demographic groups. 
This indicates that the GRPO-based reward design successfully reduces the tendency of the model to produce biased or stereotypical content, even without direct supervision from labeled bias datasets. 
The reinforcement signal provided by the DeBERTa-based discriminator encourages the model to internalize a more balanced judgment across diverse inputs.

Meanwhile, the minor change in fluency (-0.01) confirms that the optimization process does not significantly affect the model's surface-level language ability. 
The slight increase in relevance (0.81 → 0.84) implies that the model not only avoids unfair patterns but also produces more coherent and contextually grounded answers. 
From a behavioral perspective, this reflects a better trade-off between ethical awareness and response fidelity.

In summary, the GRPO fine-tuning procedure yields a model that is more consistent, fair, and contextually aligned without sacrificing expressiveness or grammatical quality. 
These findings verify that reinforcement-based reward modeling can serve as an effective lightweight alternative to full-scale supervised alignment pipelines, providing a practical path for enhancing ethical reasoning in large language models.


\section{Conclusion}

This study proposed a reinforcement-based fine-tuning approach to enhance the ethical alignment of large language models. By leveraging a composite reward that combines fairness, semantic relevance, fluency, and length control, our method effectively guides the model toward producing socially neutral yet contextually coherent responses. Experimental evaluation shows that the GRPO-tuned model significantly improves fairness metrics while maintaining naturalness and relevance, confirming the effectiveness of using DeBERTa-based discriminators as fairness evaluators.

Future work will extend this framework in several directions. First, we plan to incorporate multilingual and cross-cultural fairness evaluation to better generalize ethical alignment across languages. Second, we will explore dynamic reward scheduling based on conversation context and reinforcement learning stability. Finally, integrating human preference data and more diverse ethical taxonomies could further refine the balance between fairness and expressive freedom, contributing to more responsible and inclusive language model development.


\end{document}